***Electoral Programs of German Parties 2021: A Computational Analysis Of Their Comprehensibility and Likeability Based On 'SentiArt'***

Arthur M. Jacobs[1,2] and Annette Kinder[3]

Author Note

1) Department of Experimental and Neurocognitive Psychology, Freie Universität Berlin, Germany

2) Center for Cognitive Neuroscience Berlin (CCNB), Berlin, Germany

3) Department of Education and Psychology, FU Berlin, Germany

Correspondence: Arthur M. Jacobs

Department of Experimental and Neurocognitive Psychology, Freie Universität Berlin, Habelschwerdter Allee 45 , D-14195 Berlin, Germany.

Email: ajacobs@zedat.fu-berlin.de

**Abstract**

The electoral programs of six German parties issued before the parliamentary elections of 2021 are analyzed using state-of-the-art computational tools for quantitative narrative, topic and sentiment analysis. We compare different methods for computing the textual similarity of the programs, *Jaccard Bag similarity, Latent Semantic Analysis, doc2vec*, and *sBERT*, the representational and computational complexity increasing from the 1st to the 4th method. A new similarity measure for entire documents derived from the *Fowlkes Mallows Score* is applied to *kmeans* clustering of *sBERT* transformed sentences. Using novel indices of the readability and emotion potential of texts computed via *SentiArt* (Jacobs, 2019), our data shed light on the similarities and differences of the programs regarding their length, main ideas, comprehensibility, likeability, and semantic complexity. Among others, they reveal that the programs of the SPD and CDU have the best chances to be comprehensible *and* likeable –all other things being equal–, and they raise the important issue of which similarity measure is optimal for comparing texts such as electoral programs which necessarily share a lot of words. While such analyses can not replace qualitative analyses or a deep reading of the texts, they offer predictions that can be verified in empirical studies and may serve as a motivation for changing aspects of future electoral programs potentially making them more comprehensible and/or likeable.

# 1 Introduction

Party platforms are a favorite means for voters to inform themselves and facilitate their decisions regarding general elections such as the 2021 elections for the German parliament. Usually before elections political parties therefore issue special platforms called *electoral programs*. Such platforms typically state the main ideas and goals of a party (i.e., their programs) in a concise way demarcating it from those of its competitors. Thus, the 1$^{st}$ paragraph of the preamble of the party platform of the U.S. democrats states: *America is an idea—one that has endured and evolved through war and depression, prevailed over fascism and communism, and radiated hope to far distant corners of the earth. Americans believe that diversity is our greatest strength. That protest is among the highest forms of patriotism. That our fates and fortunes are bound to rise and fall together. That even when we fall short of our highest ideals, we never stop trying to build a more perfect union.*

Like other texts such platforms can be analysed with state-of-the-art methods of *computational stylistics* (Herrmann et al., 2021; Jacobs, 2018b), such as quantitative narrative, sentiment or topic analysis, to answer questions like:

- what are the main ideas and topics?
- which programs resemble each other most?
- which platform is the most readable and easy to comprehend?
- which program is the emotionally most positive and likeable?
- which platform has the highest semantic complexity?

These platforms can also serve as a testing ground for hypotheses of research in natural language processing/NLP, computational linguistics or *Neurocognitive Poetics* (Jacobs, 2015; Willems & Jacobs, 2016), e.g. the prediction verified in this paper that mainstream parties use an overall more positive vocabulary to attract (or not repel) voters than more extreme parties at the left or right borders of the spectrum.

**Comprehensibility and Likeability**

No matter what you read, expository texts or fiction, prose or poetry, two factors influence your decision to read on: whether you understand and whether you like what you read (e.g., Jacobs et al., 2016). The same holds for electoral programs. Although reading psychology has long known this fact, when it comes to methods and models for measuring and explaining which features of a text are most important for its comprehensibility and likeability, current research provides a controversial picture. Regarding measures of *comprehensibility*, the development of sophisticated computational methods has opened up a way to overcome the deficits of the simplistic readability measures of the past, such as the still widely used *Flesch-Kincaide Index* that focuses on the number of syllables, words and sentences in a text but neglects important features such as word concreteness or cohesion (Graesser et al., 2004). For example Graesser et al.'s (2004) *Coh-metrix* software delivers over 100 text features that can be used to estimate the ease with which texts can be read, and it can be applied to expository, prose or even poetic texts, such as Shakespeare sonnets (e.g., Jacobs et al., 2017). However, Coh-metrix only works for English and can not be used for analysing German electoral programs.

Quantifying features shaping the *likeability* of texts has proven to be even more complicated than to estimate their comprehensibility (Jacobs et al., 2016). Coh-metrix, for instance, basically quantifies text features influencing cognitive processes totally neglecting emotional ones. However, more recent tools also allow to quantify emotional and aesthetic aspects of texts (Crossley et al., 2016; Jacobs, 2015). The recent *SentiArt* tool for example was especially developed for quantifying the affective-aesthetic potential (AAP) of texts and works in several languages including German, English, French, Italian or Dutch (Jacobs, 2017, 2019; https://github.com/matinho13/SentiArt). It makes use of distributed semantic models (DSM; e.g., Mikolov et al., 2013) and computes the likelihood of more or less conscious associations between a text's words and emotional key concepts like fear or anger. *SentiArt* has shown good performance in predicting human ratings of word *valence* (i.e., how positive or negative a word is perceived), the *likeability* of stories and of fictive characters like 'Harry Potter' or real persons like Einstein (Jacobs & Kinder, 2019, 2021; Jacobs et al., 2020), or even the *literariness* of metaphors (Jacobs and Kinder, 2017, 2018). The current version of *SentiArt* goes beyond its applicability for sentiment analyses by including dozens of features influencing text comprehensibility (Jacobs, 2018b; Jacobs & Kinder, 2018), thus allowing to estimate how well readers *understand and like* texts including electoral programs. Example features used in the present analyses are: the AAP, anger score, word concreteness, or sentence similarity (see Appendix for a detailed explanation of all features).

**2 Related Work**

Automatic text or sentiment analyses have been applied to political texts –or politically potentially relevant texts such as tweets– in many studies. Examples are Barack Obama's presidential speeches (e.g., Wang, 2010), speeches of Indian politicians (e.g., Katre, 2019), or the prediction of results of the U.S. or Indonesian presidential elections based on analyses of Twitter (e.g.,

Budiharto & Meiliana, 2019). However, as far as we can tell there seems to be little –if any– scientifically published work analysing German party platforms or electoral programs, although there are publically available reports from commercial companies (e.g., [https://www.inwt-statistics.de/blog-artikel-lesen/text-mining-part-3-sentiment-analyse.html)](https://www.inwt-statistics.de/blog-artikel-lesen/text-mining-part-3-sentiment-analyse.html), weekly magazines (https://www.zeit.de/politik/deutschland/2017-08/bundestagswahl-wahlprogramme-parteien-computeranalyse/seite-2), or personal blogs (e.g., [https://sebastiansauer.github.io/textmining_AfD_01/)](https://sebastiansauer.github.io/textmining_AfD_01/).

**3 Models, Methods, Hypotheses**

Here we applied the *SentiArt* algorithm to compute numerous text features relevant for answering the above questions of interest and for testing our hypotheses regarding the reading of electoral programs. *SentiArt* implements the quantification of text features that according to the Neurocognitive Poetics Model of literary reading (NCPM; Jacobs, 2015) determine to what extent a text is more or less fluently processed by the well-known left-hemispheric reading network of the brain (e.g., Liebig et al., 2017, 2021) and subcortical or right-hemispheric circuits involved in affective and aesthetic processes (Hsu et al., 2015; Jacobs & Willems, 2019; Sylvester et al., 2021; Ziegler et al., 2018). Thus, for example, the average length, frequency of occurrence (i.e., familiarity) and concreteness or imageability of the words in a text are major factors influencing comprehensibility or reading fluency, while the average valence affects text likeability (Jacobs et al., 2016). Traditionally reading psychologists have used subjective human rating data of word familiarity, imageability or valence as predictors of reading performance. However, such an approach is problematic because of its circularity (Hofmann et al., 2018) and therefore more objective computational methods have been developed to quantify e.g. the imageability or valence of words using DSMs, thus overcoming the epistemological problem and at the same time making the costly collection of thousands of human rating data in several languages redundant (e.g., Westbury et al., 2015). *SentiArt* is such a model-guided method and its predictive validity regarding human ratings, reading behavior or brain activity has been verified in a number of empirical studies that also provide evidence for the validity of the NCPM (e.g., Altmann et al., 2012; Bohrn et al., 2013; Hsu et al., 2015; Jacobs, 2015, 2017; Jacobs & Kinder, 2019; Xue et al., 2019, 2020).

How well you understand and/or like a text depends on many factors related to the (con)text-author-reader nexus (Jacobs, 2015). Naturally, not all of these can be quantified. Thus, to predict how well readers understand and how much they like a given electoral program would benefit from knowing their reading skills and political preferences. In the present analyses –like in most others in this field– such factors remain unknown. On the other hand, the authors of electoral programs try to write their texts in a way that makes them understandable, likeable and thus potentially attractive and persuasive also for people whose reading skills and political opinions they do not know. Thus, these texts are written for an ideal average reader who does not exist in reality but represents something like the best guess or approximation of real readers. Reading psychology has always worked with such *ideal reader models* (e.g., Jacobs, 2015; Just & Carpenter, 1980) –some of which have been implemented as computer programs– and obtained very promising results in predicting human performance in a variety of reading or reading-related tasks (e.g., Hofmann & Jacobs, 2014; Legge et al., 1997; Engbert et al., 2002).

A major hypothesis we'd like to advance against this theoretical background is that mainstream programs issued by political parties in or near the midddle of the spectrum use more emotionally positive words than extreme parties at the borders of the spectrum. This is because mainstream parties aim at pleasing a majority and woo voters who shy away from aggressive or extreme expressions associated with fear, anger or disgust. In other words, they try to follow the *Pollyanna principle,* i.e., a universal human tendency to use evaluatively positive words more frequently, diversely and facilely than evaluatively negative words (Boucher & Osgood, 1969). Recent computational studies established that this principle holds for large text corpora in several languages (Dodds et al., 2015), including children and youth literature (Jacobs et al., 2020).

Thus, the electoral program of the German CDU claiming to be a people's party right in the center of the spectrum should be clearly more positive than those of the left- and right-wing parties, i.e. the LINKE and the AFD. There is some evidence for this hypothesis coming from an analysis published in the German magazine 'Die ZEIT" in 2017. It used the word-list based sentiment analysis tool 'SentimentWortschatz' (featuring a rather limited vocabulary of ~3000 words; in comparison the German version of *SentiArt* used here covers >115.000 words) to estimate the development of the 'mood' of seven electoral programs over seven election years (1990 – 2013). The results suggested that i) only CDU and SPD continuously evoke a positive mood (i.e., having more words of positive valence than of negative), and ii) all parties show a trend towards a more positive mood over the years, even the LINKE and AFD which showed the most negative mood. A second hypothesis we tested is that the applied similarity measures should produce clusters compatible with the left-to-right spectrum of the six parties considered here, e.g. group SPD with GRÜNE rather than with AFD.

## 4 Overall Statistics and Main Topics

Which party has the longest program, which uses the shortest sentences? What are main topics of each? These questions are answered in this section.

**Table 1. Global Statistics For The Six Electoral Programs**

| party | Number of Sentences | Number of Words | Word Length (number of syllables) | Sentence Length (number of words) | Number of pages in .pdf file (incl. tables, figures etc.) |
|---|---|---|---|---|---|
| **afd** | ~1500* | ~25000 | 2.47 | 18.26 | 96 |
| **cdu** | ~2500 | ~44000 | 2.41 | 17.21 | 76 |
| **fdp** | ~2000 | ~36000 | 2.5 | 17.4 | 68 |
| **grüne** | ~3500 | ~66000 | 2.44 | 18.48 | 272 |
| **linke** | ~4500 | ~69000 | 2.36 | 15.53 | 168 |
| **spd** | ~1500 | ~23000 | 2.43 | 16.79 | 66 |

* the exact number of sentences per text can vary due to the relative fuzziness of sentence parsers for German (see Appendix for details)

The data in Table 1 reveal that the left (LINKE) and green parties (GRÜNE) expect the most of their readers with the longest programs of ~4500 and ~3500 sentences, respectively. CDU, FDP, AFD, and SPD follow in that order, the SPD's program being about 3000 sentences shorter than that of the LINKE. Taking a conservative estimate for the average reading rate (~200 words per minute/wpm) as a basis, reading the entire program of the LINKE without any breaks would thus take ~6 hours [(69000/200)/60 = 5.75]. This is almost a working day and one could easily read a literary novel in that time. Even a more experienced reader with an optimistic rate of 300 wpm would still need about 4 hours to fulfill the demanding task. In contrast, readers interested in the SPD are done with their program after about 2 hours (at a rate of 200 wpm). Of course, such *ideal reader model* computations are only approximations and it is perhaps unlikely that any reader would process these programs in a single reading act. Leafing and skimming through them with occasional stops for deeper readings motivated by section headings like 'Entlastung und Faire Steuern', for example, is the more likely option. Still, these notable differences in the extent of the programs tell us something about what parties expect of their supporters and interested readers.

**Figure 1. Selection of Main Topics of the Six Electoral Programs**

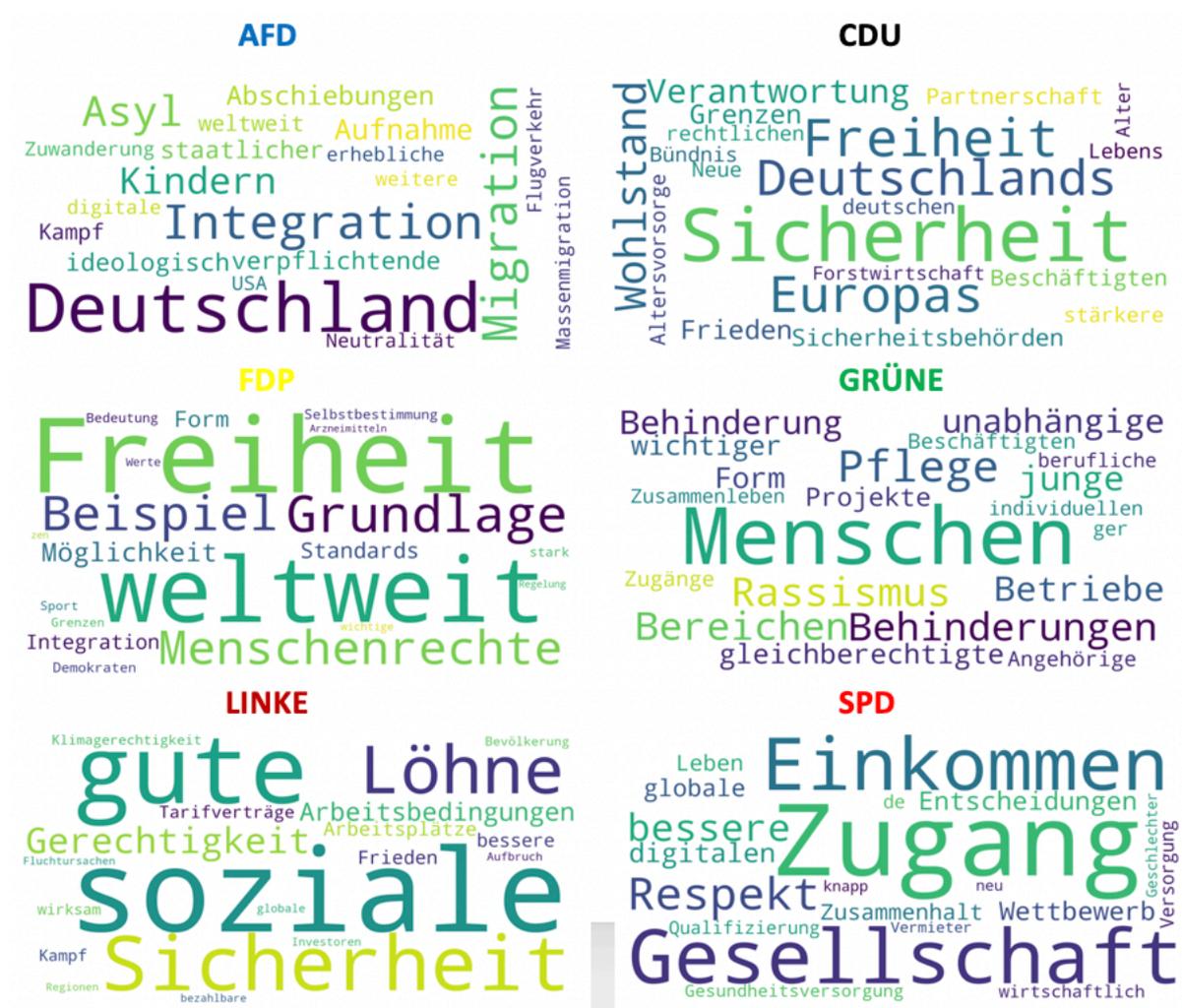

Figure 1 summarizes the main topics of each program (in alphabetical order from left to right) in the practical form of word clouds, as generated by a topic analysis using a standard library (the gensim LDAmallet; Rehurek & Sojka, 2010[1]). For simplicity, for each party, we present the most distinctive of the ten topics with the highest probability out of overall 25 computed topics. A quick assessment reveals that the AFD's topic emphasizes immigration and deportation, while the CDU's foregrounds security and liberty and prosperity. A major topic of the FDP turns around the notions of liberty and human rights, while the GRÜNE stresses a 'people topic' with key terms like housing, disability, home care or youth welfare. Finally, the LINKE deals with social security and justice, wages, jobs, or labour conditions and agreements, while a topic important for the SPD turns around society, income, access and respect. Note that the terms in such topic analyses do not automatically reveal whether a party is *for* or *against* them, something which can often but not always be deduced from the co-occurrence with other terms, though. In any case, such data based on frequency of occurrence must be interpreted with care in the light of other related data like those in section 5, for instance.

---

[1] Only nouns and adjectives were included in this topic analysis having shown the most convincing results in a pilot study.

**5 Text Similarity**

The above distinctive main topics are one side of the programs which otherwise also share a number of ideas. There is a variety of NLP methods for computing the similarity of documents which can produce notably different results, and there seems to be no gold standard so far. This is why we used four such methods for comparison, each representing a different generation and level of complexity of language models. Given that despite differences in focus electoral programs necessarily share a lot of words, the issue of which similarity measure is optimal seems crucial. The first method, *Jaccard Bag similarity,* is a simple set-theoretic measure from the beginning of the last century, based on a comparison of the unique words that are shared and not shared in two documents. We include this rather crude measure because it was the index used in the above mentioned study of 'Die ZEIT'. The 2$^{nd}$, *latent semantic analysis* (LSA) published almost a century later is based on a statistical transformation of the co-occurrence of terms in documents (LSA; Deerwester et al., 1990[2]). The 3$^{rd}$, *doc2vec*, published about another decade later is based on 'paragraph' vectors (Le & Mikolov, 2014): these are document embeddings learnt by a neural net, based on an extension of *word2vec* (Mikolov et al., 2013) to sequences of arbitrary length. A sequence can be anything from sentences to documents; here we used the LSA-transformed 2d representations of the centroids of chunks of 1000 words (van Cranenburgh et al., 2019). All three methods ignore syntax and thus lose relevant information. So as a fourth method we chose one that represents the latest generation of language models, so-called *transformers*, which take sentence structure into account: the *sBERT* algorithm[3] (Reimers & Gurevych, 2019;

---

[2] Each program was LSA-transformed into a 2d representation of the *tfidf* (term frequency inverse document frequency) of the entire document using the *gensim* library.

[3] We first transformed each sentence of each program into a 768d vector using the German 'dbmdz/bert-base-german-uncased' model (https://huggingface.co/dbmdz/bert-base-german-uncased). Since computing the cosine similarity between all sentences of all programs was computationally prohibitive, we then clustered these vectors pairwise (i.e., afd vs. cdu, afd vs. fdp etc.) using the *kmeans* algorithm (https://scikit-learn.org/stable/modules/clustering.html#k-means), with k = 2. We then computed the *Fowlkes Mallows Score* (FMS, https://scikit-learn.org/stable/modules/generated/sklearn.metrics.fowlkes_mallows_score.html#sklearn.metrics.fowlkes_mallows_score). As an index of clustering performance, it measures the similarity of two clusterings of a set of points and varies between 1 (perfect match between predicted cluster labels and ground truth labels) and 0 (total mismatch). Since we knew the ground truth labels for each sentence (i.e., the names of the parties), we could use 1-FMS as a similarity measure: a perfect sentence clustering (via kmeans) means maximum dissimilarity between the two clusters and vice versa, e.g. the AFD sentences are perfectly predicted as one cluster and the CDU sentences, too. The more mistakes *kmeans* makes (e.g., labeling an AFD sentence as CDU or vice versa), the higher the similarity between two different programs and the lower the FMS. This generated the matrix shown in Figure 2f.

https://www.sbert.net/). The representational and computational complexity increases from the 1$^{st}$ to the 4$^{th}$ method.

Figure 2 (a-f, from top to bottom and left to right) summarizes the results. The *LSA* data suggest that the programs of GRÜNE and SPD are very close, as are –somewhat astonishingly– those of FDP and AFD, while those of LINKE and CDU stand apart. The *doc2vec* data provide a similar picture: again the SPD is closest to the GRÜNE and closer to the CDU than to the FDP, which is now farther apart from the AFD. LINKE and AFD now are clearly separated from the rest supporting our 2$^{nd}$ hypothesis. An example illustration of *kmeans* clustered *doc2vec* representations is given in Figures 2c and d showing the *2d semantic overlap* between AFD and FDP or LINKE, respectively: it is clearly higher for AFD and FDP (*Fowlkes Mallows Similarity* = .487) than for AFD and LINKE (*Fowlkes Mallows Similarity* = .39; note that the overlap between AFD and FDP in the reduced 2d space of Figure 2c is clearly greater than in the full 300d space of Figure 2b).

Considering the four graphs of the lower half of Figure 2, quite different similarity patterns emerge. The *Jaccard Bag similarity* index (Figure 2e), which does not take word frequencies into account, generates rather tiny differences between the programs (note that for better visibility the data have been rounded up to two decimals). A first observation is that the AFD bears the smallest average similarity of all parties (.20), while the GRÜNE share the most words with all other parties (.234). Second, and quite amazingly, if this index were valid, the programs of FDP and SPD are more similar to the AFD's than are those of the other parties. In the 2017 analysis of the 'Die Zeit', the CDU was closest to AFD, followed by FDP and SPD. Whether this is a coincidence, a measurement fuzziness, or an intended change in the CDU's vocabulary we cannot tell. Third, the greatest similarity of all (.26) exists between the programs of the GRÜNE and the LINKE, which was already the case in the 2017 data from 'Die ZEIT". The SPD is closer to the GRÜNE (.24), FDP (.23) and CDU(.22) than to the two extreme parties (.21) which is also true for the CDU. According to the *Fowlkes Mallows Similarity* –based on a kmeans clustering of sBERT sentence similarity– the program closest to the AFD is again that of the FDP, while that closest to the FDP is the CDU's. However, the SPD is now closer to CDU and FDP than to GRÜNE and LINKE.

Running head: *German Electoral Programs 2021 Quantified*

**Figure 2a - f. LSA (a) vs. DOC2VEC (b) representations for the Six Electoral Programs; overlap between DOC2VEC representations for AFD and FDP (c) vs. AFD and LINKE (d); Jaccard Bag (e) vs. Fowlkes Mallows Score (f) Similarity for the Six Electoral Programs.**

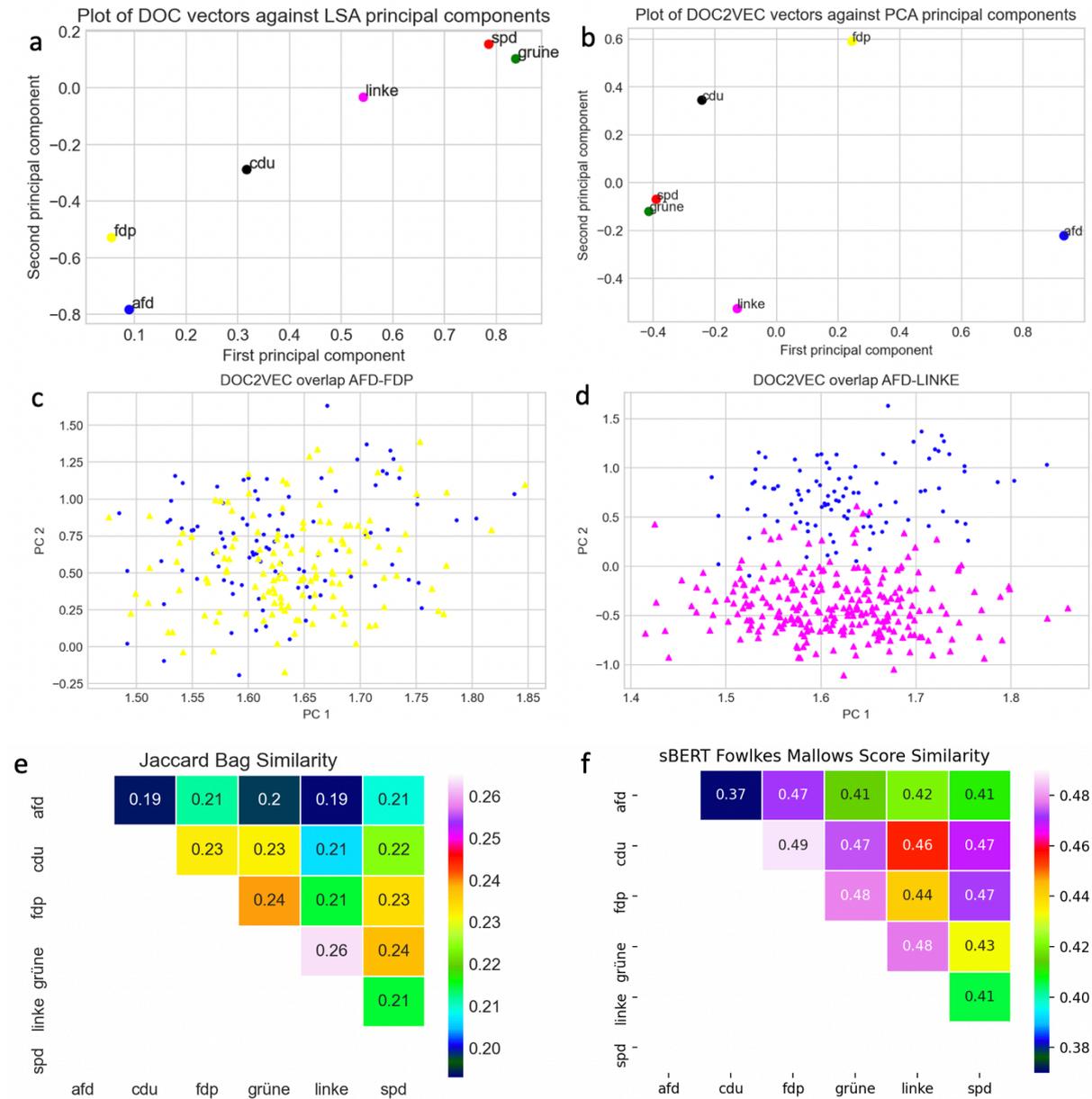

To sum up section 5, basically with different inputs (e.g., single words, text chunks, sentences) and different methods (LSA, doc2vec, Jaccard, sBERT) one obtains different output patterns. The lesson to learn from these data is that caution is demanded when picking methods of document similarity computation and clustering for electoral programs –or any other text type– and interpreting their results. Although like many others working in the fields of education, psychology or cognitive science (e.g., Lee et al., 2005), we personally also have often obtained the most convincing results with the LSA method for analyses of works of fiction and poetry, we know of no systematic study allowing to say with sufficient safety which method is optimal for the analysis of electoral programs. More computational and, foremost, empirical research is of the essence here, since we lack a ground truth or gold standard for the perceived similarities (e.g., human ratings) between the 2021 German electoral programs.

**6 Readability and Emotion Potential**

As psychologists we would like to distinguish between the terms *comprehensibility* and *likeability* on the one hand, and *readability* and *emotion potential* on the other. While the latter refer to measurable properties of texts, the former also depend on 'the eye of the beholder', i.e. take into account reader variables such as reading age and skill, or personal habits and preferences. The latter can predict the former to a certain extent, but a full-fledged psychological model of reader responses to electoral programs or other texts would involve collecting data about reader variables, too.

Instead of using standard simplified univariate indicators of the complex constructs *readability* and *emotion potential* we opted for a multivariate, factor-analytic approach similar to the one applied in Coh-metrix (Graesser et al., 2004). We first selected a set of 23 text features (out of an ensemble of >50 computed by *SentiArt*) that can directly be associated with either of the two constructs. We than ran several exploratory factor analyses varying the rotation method (e.g., varimax vs. orthomin), factoring method (maximum likelihood vs. principal axis), and prior communality (principal components vs. common factor analysis) and finally selected the one accounting for a maximum of variance: the varimax, principal axis, five factors model which accounted for 77.4% of variance.

Two of the five factors loaded highest on features related to the emotion potential. The first, called *valence*, loaded high (>.5) on five affective word features (overall affective-aesthetic potential/AAP, AAP for nouns, AAP for verbs, IMS_valence[4], ratio of positive/negative words); the second, called *arousal*, also loaded high on five affective word features (arousal, anger, disgust, fear, and sadness). The other three factors loaded highest on features related to *readability*. The first, called *word concreteness*, loaded high on two features (concreteness and imageability); the second, called *word complexity*, loaded high on four features (number of letters, number of syllables, orthographic dissimilarity score, and sonsority score); the third, called *sentence complexity*, loaded high on six features (number of words, number of content words, phrase density, sentence syllable index, content word overlap, and average sentence similarity). Note that concreteness works in the opposite direction of the two other factors which decrease readability. A detailed explanation of these features is given in Appendix B.

---

[4] IMS stands for ‚Institut für Maschinelle Sprachverarbeitung': https://www.ims.uni-stuttgart.de/ (see Appendix B).

Figure 3 sumarizes the readability of the six programs. The LINKE's program appears to have the highest readability combining the 2[nd] highest concreteness score with the lowest scores for word and sentence complexity. However, it is also the longest of all six (cf. Table 1). The programs of SPD and CDU also seem to be easier to read than those of the GRÜNE, FDP and AFD which exhibit high levels of word and sentence complexity, although the AFD's program should be easier to read due to its highest concreteness score.

**Figure 3. Readability of the Six Electoral Programs**

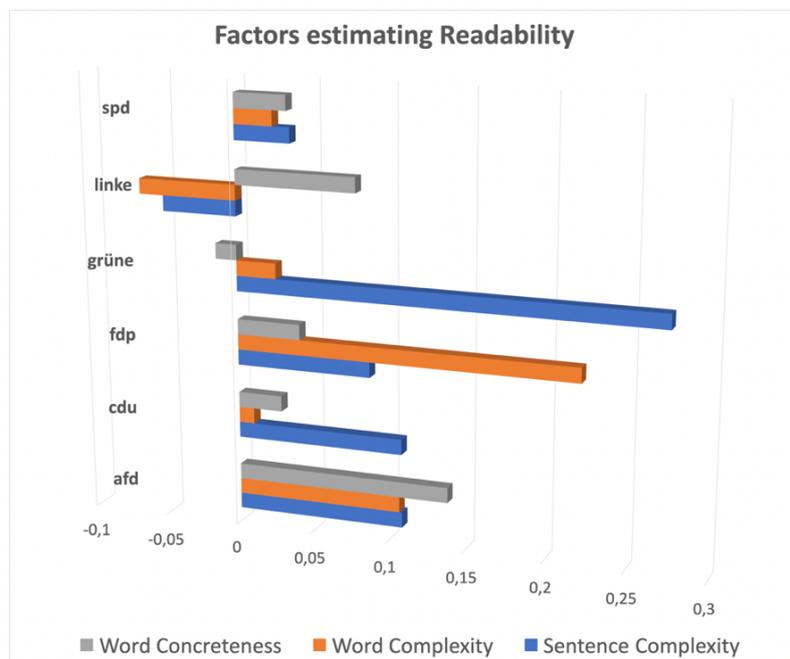

As an example of the GRÜNE's somewhat exorbitant sentence complexity consider the following 46-word sentence behemoth: ,*Dies erfordert eine grundsätzliche Änderung der bisherigen standardisierten Bewertungsverfahren, Berechnungsgrundlagen und Kriterien unter Berücksichtigung der tatsächlichen Klima und Umweltkosten, die gründliche Prüfung von Alternativen, die auch andere Verkehrsträger einbezieht, eine Verbesserung der bisher unzureichenden Beteiligung der Bürgerinnen und Verbände sowie die Abkehr vom sogenannten Finanzierungskreislauf Straße.*' Less skilled or dyslexic readers are more than challenged by such verbal accumulations, but even normally skilled readers will have problems due to verbal working memory limitations. Assuming a capacity of five words –an estimate of the average reading span–, when a reader's gaze arrives at the first word after the 2[nd] comma (*die*), about ¾ of the previous words will already have faded from working memory thus making a backward saccade (i.e., an eye movement to the left) likely in order to refresh the information necessary to make sense out of the sentence. Another example from the AFD is not less challenging: '*Im Einzelnen wollen wir die Verfahren beschleunigen und dafür (a) die noch aus dem 19. Jahrhundert stammende Prozessordnung modernisieren, (b) mehr Personal für das Justizwesen einstellen, (c) Schwerkriminalität durch Erleichterung der Inhaftnahme wirkungsvoller bekämpfen, (d) weniger Strafen zur Bewährung aussetzen und (e) das Mindeststrafmaß bei „Messerdelikten" erhöhen.*' A reading psychologist might wonder what the authors of such verbal productions think about their readers. If they encounter too many of this kind they may well start liking the text less than their authors hoped.

Now let's consider the other factor that influences how much readers like a text. In Figure 4 we take a look at which programs potentially evoke the most positive mood in their readers and the answer is quite clear based on clear differences in the factor 'valence' (the second factor, arousal, is ignored here due to its negligeable variance: CDU and SPD. In contrast, AFD and LINKE risk to evoke rather negative moods in their readers perhaps betting on the idea that this negative mood will somehow be converted into positive votes and/or political radicalization. GRÜNE and FDP also offer positively biased texts to their readers, again showing that mainstream and extreme parties differ notably in their verbal expressions. These data correspond well with those of 'Die ZEIT' from 2017. In the light of these data one can expect readers of the GRÜNE's program to forgive the author's their occasional sentence behemoths since the text is overall very positive. For the AFD things are different though: complex sentences paired with negative valence is not a cocktail that makes readers happy, at least not when they aren't party supporters. To summarize, if our computational indices were valid, then –all other things being equal– the SPD's and CDU's programs would have the best chances to be comprehensible *and* likeable. But, of course, other factors like a reader's personality and preferences also play a role which can not be assessed with such tools.

**Figure 4. Emotion Potential of the Six Electoral Programs**

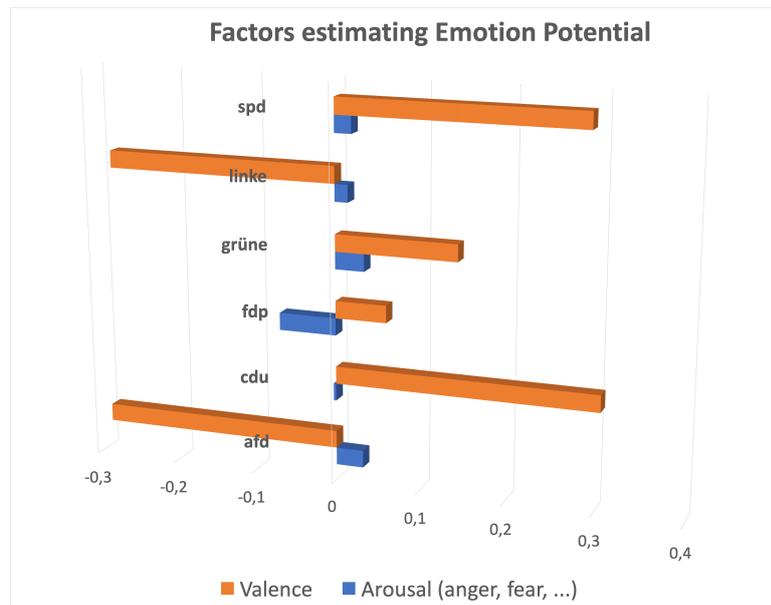

## 7 Semantic Complexity

In dynamic interaction with syntactic complexity (see Figure 3) semantic complexity also plays a role for the comprehensibility and likeability of texts. Two recent measures of semantic complexity that showed promising results, e.g. for the prediction of the literariness of Dutch novels (van Cranenburgh et al., 2019), are the *intra-textual variance* (ITV; see equations in Appendix) and *stepwise distance* (SWD). According to the data in Figure 5 (left panel) below the LINKE's program is the most semantically complex (i.e., having the highest values of ITV and SDW), but it also had the lowest level of sentence complexity. The CDU's and FDP's programs are the least semantically complex. Those of the other three parties all are closer to the LINKE's level of semantic complexity than to those of CDU and FDP. The right panel of Figure 5 contrasts the ITV values for text chunks from the LINKE (magenta) and CDU (black) programs in a 2d space generated via a *principal component analysis*. The ITV corresponds to the mean of the squared distances between the individual text chunks (dots) and the centroids (i.e., means of chunk vectors marked by star and triangle). As is also apparent from the left panel of Figure 5, this is greater for the LINKE than for the CDU (i.e., a larger dispersion of the dots).

**Figure 5. ITV and SWD (left) of the Six Electoral Programs, ITV for two Programs**

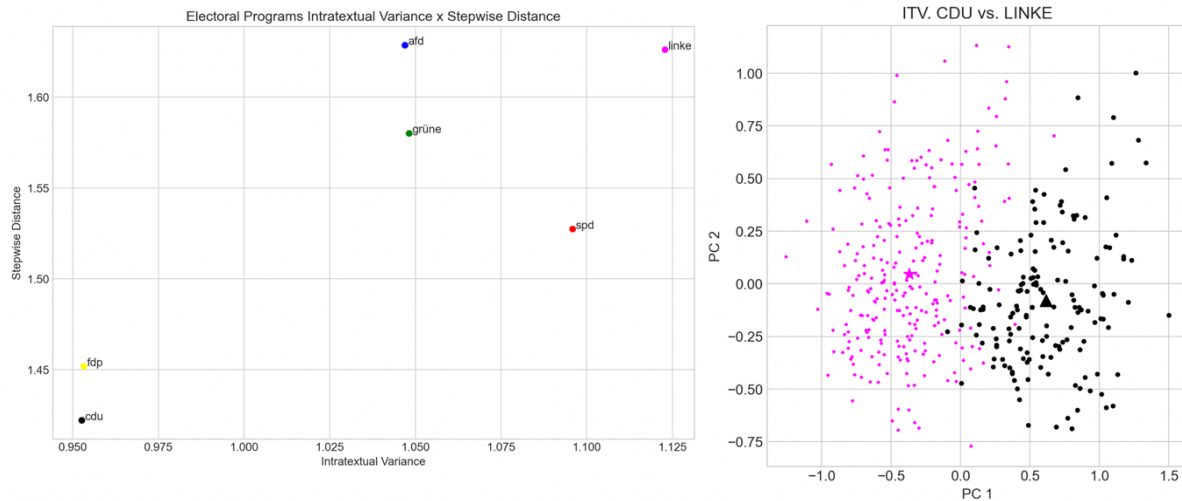

**8 Discussion, Limitations and Outlook**

The results of multiple computational analyses of the present six electoral programs shed light on their textual similarities and differences regarding text length, main topics, readability, emotion potential, and semantic complexity. Overall, the biggest differences appear between the programs of the extreme parties (AFD and LINKE) and those of the mainstream parties (CDU, SPD, GRÜNE, FDP). Despite obvious differences in selected topics (Figure 1), measures of the overall lexico-semantic and sentence similarity (Figure 2) revealed a mixed picture. For example, while the *LSA, doc2vec,* and *Jaccard* methods suggest that SPD and GRÜNE have similar programs, the one measure based on sentence similarities (*Fowlkes Mallows Similarity)* puts the SPD nearer to the CDU and FDP than to the GRÜNE. This raises the important issues of which input grain size (e.g., single words, sentences, chunks), which similarity measure, and which dimensionality or reduction method are optimal for such purposes and why. We can only conclude that more research is needed to find out the best methods for this field and that until that happens one should always compare results of several methods before any premature interpretation leads to potential fallacies.

Regarding text comprehensibility and likeability, LINKE, SPD and CDU offer programs that are relatively easy to read, while GRÜNE, FDP and AFD challenge readers with relatively high levels of word and sentence complexity. As has been shown previously and in agreement with our hypothesis, in contrast to the extreme parties all mainstream parties try to 'make their readers happy' by using an overall positively toned vocabulary. Finally, FDP and CDU produced texts of lower semantic complexity than the other parties. Thus, an average politically unbiased, maximally 'objective' reader who took her time to read all six programs (taking several days) would probably like the SPD's and CDU's very comprehensible and positive programs best (all other things being equal). Of course, this is a speculation; but it also is a testable prediction based on scientific methods and thus can be empirically verified in future studies that at least some political parties may be interested to run.

The present computational analyses are based on standard NLP methods thus inheriting their strenghts and weaknesses. One obvious limitation regarding the sentiment analyses, for exmaple, is that they are based on indiviual words and thus the sentiment of a sentence is computed as the simple average of the sentiment values of its content words. Although this is the most widely used standard procedure, future work might consider using aspect-based sentiment analysis (ABSA) that can deal with forms of negations and other features modifying the sentiment of larger semantic units. However, although our *SentiArt* tool does not include aspect-based analyses it predicted human liking ratings of entire text passages of a prototypical uncanny story very well (Jacobs & Kinder, 2019). Furthermore, our novel multivariate measures of the readability and emotion potential of texts still require empirical validation and thus should be considered as exploratory. The same holds for the novel application of the *Fowlkes Mallows Score* to the computation of document similarities based on a *kmeans* clustering of *sBERT* transformed sentences.

In any case, that such computational approaches like the present can at best complement a qualitative analysis, deep reading, and content-related debate is clear. However, they raise important issues –the question of which method is optimal for computing the similarity between electoral programs not being the least–, offer predictions that can be verified and thus can perhaps even serve as a motivation for changing aspects of future electoral programs potentially making them more comprehensible and/or likeable.

Running head: *German Electoral Programs 2021 Quantified*

**Appendix**

**A. Text Preprocessing**

All texts were extracted from the .pdf files publically available at: https://www.bundestagswahl-2021.de/wahlprogramme/ and transformed to .txt files. They were then pre-processed using standard NLP python routines available from the open access NLTK library (https://www.nltk.org/) and others. In particular, every sentence in each text was parsed, tokenized and POS-tagged using the *treetagger* (https://www.cis.uni-muenchen.de/~schmid/tools/TreeTagger/) or *SoMaJo* routines, the latter being optimised for German (https://github.com/tsproisl/SoMaJo). Using the *SentiArt* tool each word in each sentence of each text was quantified in terms of a number of features such as word length, concreteness, AAP etc. and feature values were then aggregated across sentences and texts. For example, the AAP (overall) feature for a given text is the mean of all content words' AAP values of that text, the AAPn feature is the mean of all nouns' AAP values of that text and so on. The AAP values for a given word in a given sentence computed here can directly by retrieved from the *SentiArt*.xlxs table available on github (https://github.com/matinho13/SentiArt). This table provides values for ~115k words covering ~55% of the words appearing in the present texts.

**B. Computation of features for readability and emotion potential**

The features were computed as follows via *SentiArt*.

**1. Readability**

**Factor Concreteness**

1.1 Concreteness. The mean of the IMS concreteness values for the words in a sentence.

1.2 Imageability. The mean of the IMS imageability values for the words in a sentence. These values based on a machine learning method applied to human ratings from different sources are available at: https://www.ims.uni-stuttgart.de/en/research/resources/experiment-data/affective-norms/ (Köper & Schulte im Walde, 2016).

**Factor Word Complexity**

1.3 Word Length. The mean number of letters per word in a sentence.

1.4 Syllables. The mean number of syllables per word in a sentence.

1.5 Orthographic Dissimilarity Score. The mean Levenshtein distance between the words in a sentence and all other words in the reference corpus (SDEWAC).

1.6 Sonority Score. The sum of the phonemes' sonority hierarchy values (cf. Vennemann, 1988) of a word divided by the square root of its length in letters (Jacobs, 2017, 2918b).

Longer words and words with a lower orthographic dissimilarity or higher sonority score attract longer total reading times (Xu et al., 2019).

**Factor Sentence Complexity**

1.7 Sentence Length. Mean number of words per sentence.

1.8 Number of content words. The sum of the number of nouns, verbs, adjectives and adverbs per sentence, as determined by *treetagger*.

1.9 Phrase Density. The number of phrases per sentence as determined by the *pattern* library (https://github.com/clips).

1.10 Sentence Syllable Index (SSI). The product of the number of tokens and mean number of syllables per sentence, the latter being computed via the *syllapy* library (https://github.com/mholtzscher/syllapy). Larger SSI values indicate that a sentence is theoretically less readable resulting in longer reading times (cf. Jacobs & Kinder, 2020).

1.11 Content Word Overlap. The number of content words that are shared by two consecutive sentences. This is a measure of the cohesion of a text, l (e.g., Graesser et al., 2004).

1.12 Sentence Similarity. The mean of the semantic similarity between two consecutive sentences, as measured via the sBERT model (Reimers & Gurevych, 2019).

**2. Emotion Potential**

**Factor Valence**

2.1 AAP. The mean of the AAP values for the words in a sentence. AAP is the average semantic similarity between a word in the text and *m* positive labels (*lpos*_1-60, = affection, amuse, …, unity) minus the average similarity between each word and *n* negative labels (*lneg*_1-60, = abominable, …, ugly; see equ. 1). The 120 labels were those published in earlier papers (e.g., Jacobs, 2017, 2018b). The semantic similarity between pairs of words (i.e., a target and a lable) is computed using a vector space model created by the authors by applying the *word2vec* algorithm (Mikolov et al., 2013) to the *sdewac* corpus (Sentence Dewac; https://www.ims.uni-stuttgart.de/en/). The final model contained a 300d *skipgram* vector for each of ~1.6 million words (for further details see Jacobs & Kinder, 2019). The cosine of two word vectors gives their similarity value. Higher AAP values indicate a text's higher potential for evoking positive affective responses including aesthetic feelings of 'liking' and 'beauty' (Jacobs, 2017).

$$v(w) = \sum_{i=1}^{m} s(w, l_{pos_i})/m - \sum_{i=1}^{n} s(w, l_{neg_i})/n \qquad \text{equ. (1)}$$

2.2 IMS_valence. The mean of the IMS valence values for the words in a sentence.

2.3 PNR. The ratio of positive to negative words in a sentence as determinedd by their AAP values.

**Factor Arousal**

2.4 Arousal. The mean of the semantic relatedness/similarity values between each word in a sentence and the labels successfully used by Westbury et al. (2015) to predict human arousal ratings.

2.5 Anger, Disgust, Fear, and Sadness. The mean of the semantic relatedness values between each word in a sentence and the labels 'anger', 'disgust', 'fear', and 'sadness', respectively. Within the context of NLP sentiment analyses, a text's overall ANGER or DISGUST scores for instance indicate that the text has a higher theoretical probability to induce arousing emotional responses associated with being angry/annoyed or offended/disgusted in readers.